# ADVERSARIALLY ROBUST DEEPFAKE MEDIA DETECTION USING FUSED CONVOLUTIONAL NEURAL NETWORK PREDICTIONS

*Sohail Ahmed Khan*[⋆†]    *Dr. Alessandro Artusi*[⋆]    *Dr. Hang Dai*[†]

[⋆] DeepCamera MRG, CYENS Centre of Excellence, Nicosia, Cyprus [†] Computer Vision Lab, MBZUAI, Abu Dhabi, UAE

## ABSTRACT

Deepfakes are synthetically generated images, videos or audios, which fraudsters use to manipulate legitimate information. Current deepfake detection systems struggle against unseen data. To address this, we employ three different deep Convolutional Neural Network (CNN) models, (1) VGG16, (2) InceptionV3, and (3) XceptionNet to classify fake and real images extracted from videos. We also constructed a fusion of the deep CNN models to improve the robustness and generalisation capability. The proposed technique outperforms state-of-the-art models with 96.5% accuracy, when tested on publicly available DeepFake Detection Challenge (DFDC) test data, comprising of 400 videos. The fusion model achieves 99% accuracy on lower quality DeepFakeTIMIT dataset videos and 91.88% on higher quality DeepFakeTIMIT videos. In addition to this, we prove that prediction fusion is more robust against adversarial attacks. If one model is compromised by an adversarial attack, the prediction fusion does not let it affect the overall classification.

## 1. INTRODUCTION

Latest progress in the fields of computer vision, machine learning has made it significantly effortless to generate trustworthy fake media e.g. video, image and audio. Generative adversarial networks (GANs) [1] are being employed to generate highly realistic images as well as videos, of anyone saying and doing anything that its creator wishes. This synthesized content is known as deepfakes. Deepfake videos can sometimes be very entertaining, but they can also be employed for malicious purposes.

In a GAN architecture, two neural networks, namely, (a) Generator and (b) Discriminator, confront each other. The generator network trains on a data set and then generates a forged video, while the discriminator attempts to detect the forgeries. The generator generates fakes until the discriminator can't detect the forgery.

The current deepfake detection approaches struggle against unseen data i.e. poor generalization capability [2, 3]. Studies suggest that when the detection models are trained videos generated using specific type of deepfake generation technique [4] e.g. faceswap, facial attribute manipulation, face synthesis, they tend to perform poorly on videos generated using different manipulation techniques. Previously proposed approaches do not significantly focus on the importance of image pre-processing, i.e. image augmentations, which can be helpful in the task of deepfake detection [3].

To solve this issue, we are proposing an holistic as well as a simple approach where the performances of well-know deep-learning architectures are averaged. The single architecture models as well as the proposed approach are first trained on DFDC [5] dataset and then tested on the DeepFakeTIMIT dataset [6]. The proposed strategy is capable to significantly improve the generalisation performances of the single deeplearning architecture. We also employ heavy image augmentations to make our training data more diverse, so that the model can learn diverse set of features.

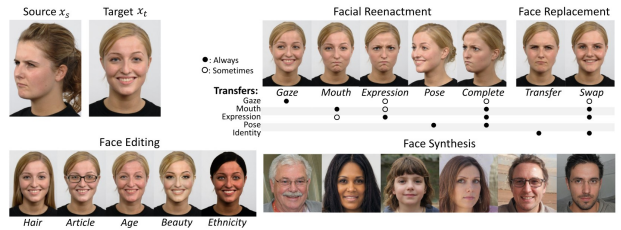

Fig. 1. DeepFake generation techniques [4].

We attack one of our models in the ensemble using Fast Sign Gradient Method (FGSM) Attack[7]. In FGSM attack, the gradients of the neural network are used to create an adversarial example. For any given image, the FGSM method generates a new image, while trying to maximise the loss using the gradients of the original input image. The new image is called the adversarial image. This process can be written as follows:

$$adv_x = x + \epsilon * sign(\nabla_x J(\theta, x, y))  \quad (1)$$

We attack one of our model using FGSM attack. Individually, our model makes a wrong classification on adversarial image. But when we fuse the predictions of all 3 models together, the adversarial image does not affect the classification result.

## 2. RELATED WORK

In most of the recent works, authors employed CNN based architectures along with other strategies to detect deepfake images/videos.

Xuan et al. in [2] used image augmentation techniques such as, Gaussian blur and Gaussian noise, during the image preprocessing step, to remove low level high frequency artifacts of GAN images. This approach increases the pixel level similarity between real images and fake images making the classifier to learn more important features.

Researchers in [8] proposed a recurrent convolutional model (RCN). A convolutional network DenseNet and the gated recurrent neural network was integrated to exploit temporal discrepancies across different frames of a video. The proposed method is tested on the FaceForensics++ data set, which includes 1,000 videos, and shows promising results.

Similarly, Guera and Delp in their study [9] proposed a pipeline based on CNN and long short term memory (LSTM) networks to detect deepfake videos. They found that deepfake videos posses intra-frame temporal inconsistencies between frames. The deep CNN extracts frame-level features, which are then given to the LSTM to create a temporal sequence descriptor. A fully-connected layer is finally employed to classify synthesized videos.

Nguyen et al. in [10] proposed capsule networks based model to detect synthesized images/ videos. The proposed model is evaluated using four datasets comprising of a wide range of forged images and videos. Proposed method achieves the best performance compared to its competing methods in all of these data sets.

Again, Nguyen et al. in [11] proposed a convolutional neural network based architecture which employs multi-task learning approach to detect manipulated images/videos and also, to segment the doctored regions. The network comprises of an encoder and a Y-shaped decoder. They evaluated the model using FaceForensics and FaceForensics++ datasets and found that their model can quickly learn to deal with unseen data by looking at only a few samples for fine-tuning.

Researchers in [12] proposed a novel approach to detect synthetic content using biological signals hidden in portrait videos. Researchers generate novel signal maps and employ a CNN to improve the traditional classifier for detecting synthetic content. The resulting model was tested on a number of different datasets, which achieved 96%, 94.65%, 91.50%, and 91.07% accuracy, on Face Forensics, Face Forensics++, CelebDF, and on Deep Fakes Dataset respectively.

In [13] researchers proposed a deep learning model based on two networks, both comprising a small number of layers to pinpoint the mesoscopic features of images. The model was evaluated on both an existing dataset and a dataset collected by researchers from online videos. The model achieves a very promising detection rate with more than 98% for Deepfake and 95% for Face2Face datasets respectively.

## 3. METHODOLOGY

### 3.1. Dataset

We employ DFDC dataset [5], available through Kaggle, to train our models. We train our models on only a subset of DFDC dataset i.e. around 25%. We also incorporate 1 face frame from each 95000 DFDC videos, which gives us 95000 face images. Altogether, we trained our models on around 270000 images.

We do data oversampling for real face images (i.e. duplicate real face images) to balance training data, because real images are less than fake images in quantity and the model might overfit the majority class.

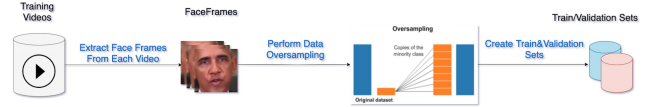

Fig. 2. Organize and balance data [4].

### 3.2. Face Detection and Image Augmentations

We employed MTCNN library to detect and extract faces from video frames [14].

Since we know from previously proposed studies [2, 3] that, the deepfake detection algorithms have a major problem of overfitting (i.e. poor generalization capability), we incorporated a number of different image augmentations. We employed ImgAug library [15] for image augmentation. We used several different augmentations to augment our data e.g. (1) image cropping, (2) random horizontal flips, (3) gaussian blur, (4) gaussian noise, (5) linear contrast and multiply (multiplies all pixels in an image with a specific value; makes the image darker or brighter).

### 3.3. Model Training

Three different CNN architectures, (1) InceptionV3 [16], (2) VGG16 [17] and (3) XceptionNet [18] are trained for the deepfake video detection task. All of the classifiers are, finetuned to classify fake and real videos. We trained the last 47 layers of the InceptionV3, last 8 layers ofVGG16, and last 35 layers of Xception, including the 4 layers listed above and froze other remaining layers.

We trained our models for different amount of epochs, 25, 50, 75, 95, and found that the models trained for 25 epochs to be performing better, and are more robust against unseen data. This is expected as training the model for higher number of epochs makes the model to overfit the training data. And we proved this during our training. We also found that training the model on all of the data at once gives better performance than the model trained on chunks of data iteratively. We employed KERAS [19] deep learning library to implement our solution.

3.4. CNN Predictions Fusion

For fusion, we take individual predictions from all of the 3 classifiers VGG16, InceptionV3, XceptionNet and averages the frame-by-frame predictions made by these CNNs on each video. At the end, the bigger probability is assigned to the video, either "real" or "fake". Figure 3 depicts this process.

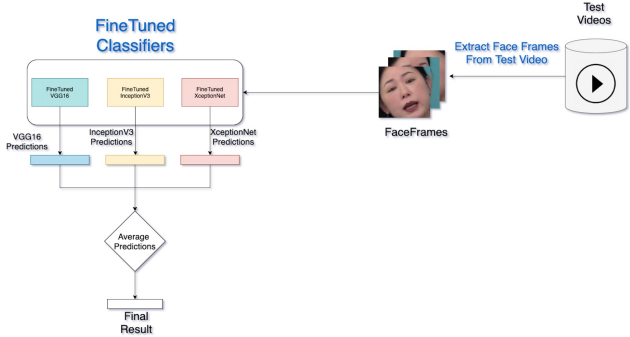

Fig. 3. CNN Prediction Fusion on Test Set.

3.5. Adversarial Attack

DeepFake solutions should be resilient to external attack, i.e., adversarial attack, and being able to rightly perform their classification task when the image/video has been distorted. To prove the resiliency capability of the proposed solution we have distorted the tested images/videos with a well known FGSM attack [7].

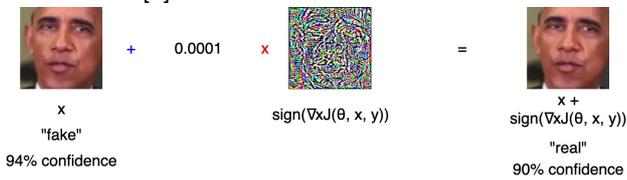

Fig. 4. FGSM Attack on VGG16 Model [7]

From figure 4 we see that a very small amount of perturbations (i.e., imperceptible to even a human eye) can cause the VGG16 model to miss-classify a given image.

[1] DFDC Leaderboard : https://www.kaggle.com/c/deepfake-detectionchallenge/leaderboard

To solve this problem, one could use a denoising approach to remove the introduced noise into the image/video. However, this requires to have available a robust denoising approach that can work for different varieties of noises. Moreover, this is requiring an extra level of know-how and complexity than the one requiring for the development of a DeepFake technology. To overcome these drawbacks, we have tested it on the proposed solution, the fused prediction, which is capable to correctly classify the noised image/video. This is mainly due to the resiliency of the some of the models used in the fused proposed approach.

4. RESULTS

4.1. Kaggle DeepFake Dataset

For the evaluation, we extract faces from test videos and save in separate folders. The model makes predictions on each image (e.g. probability of image being real or fake). The predicted probabilities for each frame of a video are averaged and compared. The bigger probability, either real or fake, becomes the model's prediction for the whole video.

There are 400 videos in the DFDC public test set. Each of our models achieved more than 96% accuracy scores on the test set, with miss-classifying only a small amount of videos. We calculated the Log-Loss scores for each model. This is to compare our models with other models submitted to Kaggle for the DFDC [5] challenge. All of the 3 of our models outperformed the Kaggle models on LogLoss score, placing our solution on top of Kaggle's DFDC public leaderboard [1]. Unfortunately, we can't submit our models to Kaggle for evaluation because the competition is closed for further new submissions.

| Classifier | Accuracy | LogLoss |
|---|---|---|
| Ours (VGG16) | 96.75% | 0.13482 |
| Ours (InceptionV3) | 96.25% | 0.12077 |
| Ours (XceptionNet) | 96.25% | 0.10123 |
| Fused Predictions (VGG16 + InceptionV3 + XceptionNet) | 96.50% | 0.11140 |
| DFDC Winning Model | | 0.19207 |

Table 1. Performance Comparison on DFDC Dataset.

4.2. DeepFakeTIMIT Dataset

DeepfakeTIMIT [6] is a database of videos where faces are swapped using the open source GAN-based approach. The DeepFakeTIMIT Dataset contains 640 fake videos. 320 higher resolution videos(128x128). The remaining 320 fake videos are lower resolution videos (64x64). In DeepFakeTIMIT Dataset, there are 16 similar looking pairs of people; total 32 subjects 10 videos per person, in lower and higher resolution each. VGG16, InceptionV3, XceptionNet and afusion of 3 models (VGG16 + InceptionV3 + XceptionNet) are tested on these 640 fake videos.

| Classifier | Low Res Videos | High Res Videos |
|---|---|---|
| VGG16 | 99.06% | 82.50% |
| InceptionV3 | 96.56% | 81.56% |
| XceptionNet | 98.75% | 84.06% |
| Prediction Fusion (VGG16 + InceptionV3 + XceptionNet) | 99.68% | 91.88% |

Table 2. Model Performance on DeepFakeTIMIT Dataset.

InceptionV3 model achieved 96.56% accuracy on the lower resolution 320 videos and only miss-classified 11 videos. VGG16 achieved an accuracy of 99.06%., whereas, XceptionNet achieved 98.75% accuracy on lower resolution DeepFakeTIMIT videos. The prediction fusion strategy outperforms these individual models by scoring 99.68% accuracy.

Our experiments suggest that the deepfake detection systems struggle against higher resolution videos. All of the individual models achieved a little over 80% accuracy on the higher resolution videos, whereas the prediction fusion strategy achieved 91.88% accuracy.

## 5. DISCUSSION

A lots of studies have been conducted on the topic of deepfake detection, but we are still in the evolutionary stages to solve this problem. Almost all of the studies achieve excellent performance scores when the models are tested on the same datasets they are trained on, but when the models are tested on "in the wild" data, they seem to struggle [3]. One example is the Facebook's DFDC [5] challenge, where the winning model acheived 82.56% accuracy on the public test set, but, when tested on "in the wild" data, it achieved only 65% accuracy. Ironically, none of the participating models achieved over 70% accuracy on the private test set which contained "in the wild" videos.

Our study also confirmed that the current CNN classifiers struggle against higher resolution videos/images. Below figures show two same videos in higher and lower resolution. The model correctly classifies the lower resolution video, but fails to classify the higher resolution video correctly.

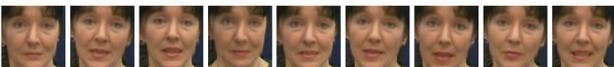

Fig. 5. Lower resolution video correctly classified as fake.

We did not trained our models on DeepFakeTIMIT dataset [6], yet, they achieved excellent results. We compare our results with the two latest studies which test their models on DFDC and DeepFakeTIMIT datasets. The results can be seen in the table 3. Our models also beat the Kaggle DFDC competition's winning model on LogLoss score as well as achieves higher accuracy then the previously proposed studies [20, 21].

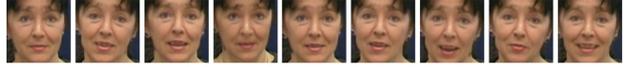

Fig. 6. Higher resolution video incorrectly classified as real.

| Study | DFDC | DeepFakeTIMIT |
|---|---|---|
| Mittal et al. [21] | 84.4% | LQ 96.30%, HQ 94.90% |
| Agarwal et al. [20] | 95.00% | N/A |
| Ours (Prediction Fusion) | 96.50% | LQ 99.68%, HQ 91.88% |

Table 3. Performance Comparison.

Our study also stresses on the importance of adversarial training to cope with adversarial attacks. We have seen during experiments that an excellent performing model is still vulnerable to adversarial attacks.

## 6. CONCLUSION AND FUTURE WORK

We propose a very simple, yet, effective method for deepfake media detection. Our model without being trained on DeepFakeTIMIT dataset, achieved excellent results, which means that our model has the generalization capability. Our model also beats other state-of-the-art systems on DFDC dataset by achieving 96.50% accuracy. However, our model seemed to struggle against higher resolution deepfake videos from the DeepFakeTIMIT dataset. In the future, we would try to address this problem. Also, to strengthen the fact the our model is good at generalization, we will test it on other deepfake detection benchmarks in the future.

We also suggest that instead of considering deepfake detection as a binary classification task, we should regard it as multi-class classification task. This is because the deepfake videos are generated using different methods, such as, face swap, face reenactment, face manipulation or entire face synthesis. Training models on these type of data, and making it classify the videos/images into binary classes does not seem to work, and thus, the models do not perform as expected.

In future, we suggest to employ multi modality data, i.e. images and audios to train the models, which might help models detect deepfakes with even better accuracy. Also, adversarial training must be in-place in order to deal with any possible adversarial attacks.

## 7. REFERENCES

[1] Ian Goodfellow, Jean Pouget-Abadie, Mehdi Mirza, Bing Xu, David Warde-Farley, Sherjil Ozair, Aaron Courville, and Yoshua Bengio, "Generative adversarial nets," in *Advances in Neural Information Processing Systems 27*, pp. 2672–2680. Curran Associates, Inc., 2014.

[2] Xinsheng Xuan, Bo Peng, Wei Wang, and Jing Dong, "On the generalization of gan image forensics," in *In Sun Z., He R., Feng J., Shan S., Guo Z. (eds) Biometric Recognition. CCBR 2019*. Springer, Cham, 2019.

[3] Polychronis Charitidis, Giorgos Kordopatis-Zilos, Symeon Papadopoulos, and Ioannis Kompatsiaris, "Investigating the impact of pre-processing and prediction aggregation on the deepfake detection task," Available: https://arxiv.org/abs/2006.07084, 2020.

[4] Yisroel Mirsky and Wenke Lee, "The creation and detection of deepfakes: A survey," in *Association for Computing Machinery (ACM)*. January, 2021.

[5] Brian Dolhansky, Joanna Bitton, Ben Pflaum, Jikuo Lu, Russ Howes, Menglin Wang, and Cristian Canton Ferrer, "The deepfake detection challenge (dfdc) dataset," Available: https://arxiv.org/abs/2006.07397, 2020.

[6] Pavel Korshunov and Sebastien Marcel, "Deepfakes: a new threat to face recognition? assessment and detection," Available: https://arxiv.org/abs/1812.08685, 2018.

[7] Ian Goodfellow, Jonathon Shlens, and Christian Szegedy, "Explaining and harnessing adversarial examples," in *International Conference on Learning Representations (ICLR)*. 2015.

[8] Ekraam Sabir, Jiaxin Cheng, Ayush Jaiswal, Wael AbdAlmageed, Iacopo Masi, and Prem Natarajan, "Recurrent convolutional strategies for face manipulation detection in videos," in *Proceedings of the IEEE/CVF Conference on Computer Vision and Pattern Recognition (CVPR) Workshops*, pp. 80–87. IEEE, 2019.

[9] David Guera and Edward J. Delp, "Deepfake video de-¨ tection using recurrent neural networks," in *15th IEEE International Conference on Advanced Video and Signal Based Surveillance (AVSS)*. IEEE, 2018.

[10] Huy H. Nguyen, Junichi Yamagishi, and Isao Echizen, "Capsule-forensics: Using capsule networks to detect forged images and videos," in *IEEE International Conference on Acoustics, Speech and Signal Processing (ICASSP)*. IEEE, 2019.

[11] Huy H. Nguyen, Fuming Fang, Junichi Yamagishi, and Isao Echizen, "Multi-task learning for detecting and segmenting manipulated facial images and videos," in *IEEE 10th International Conference on Biometrics Theory, Applications and Systems (BTAS)*. IEEE, 2019.

[12] Umur Aybars Ciftci, ˙Ilke Demir, and Lijun Yin, "Fakecatcher: Detection of synthetic portrait videos using biological signals," in *IEEE Transactions on Pattern Analysis and Machine Intelligence*. IEEE, 2020.

[13] Darius Afchar, Vincent Nozick, Junichi Yamagishi, and Isao Echizen, "Mesonet: a compact facial video forgery detection network," in *IEEE International Workshop on Information Forensics and Security (WIFS)*. IEEE, 2018.

[14] Kaipeng Zhang, Zhanpeng Zhang, Zhifeng Li, and Yu Qiao, "Joint face detection and alignment using multi-task cascaded convolutional networks," in *IEEE Signal Processing Letters*, pp. 1499–1503. IEEE, 2016.

[15] Alexander B. Jung, Kentaro Wada, Jon Crall, Satoshi Tanaka, et al., "ImgAug," Available: https://github.com/aleju/imgaug, 2020.

[16] Christian Szegedy, Vincent Vanhoucke, Sergey Ioffe, Jonathon Shlens, and Zbigniew Wojna, "Rethinking the inception architecture for computer vision," in *IEEE Conference on Computer Vision and Pattern Recognition (CVPR)*. IEEE, 2016.

[17] Karen Simonyan and Andrew Zisserman, "Joint face detection and alignment using multi-task cascaded convolutional networks," in *3rd International Conference on Learning Representations, ICLR*. 2014.

[18] Franc¸ois Chollet, "Xception: Deep learning with depthwise separable convolutions," in *IEEE Conference on Computer Vision and Pattern Recognition (CVPR)*. IEEE, 2017.

[19] Francois Chollet et al., "Keras," Available: https://github.com/fchollet/keras, 2015.

[20] Shruti Agarwal, Tarek El-Gaaly, Hany Farid, and Ser-Nam Lim, "Detecting deep-fake videos from appearance and

behavior," Available: https://arxiv.org/abs/2004.14491, 2020.

[21] Trisha Mittal, Uttaran Bhattacharya, Rohan Chandra, Aniket Bera, and Dinesh Manocha, "Emotions don't lie: An audio-visual deepfake detection method using affective cues," in *Proceedings of the 28th ACM International Conference on Multimedia*, p. 2823–2832. ACM, 2020.